\documentclass[runningheads]{llncs}
\usepackage{graphicx}
\usepackage{color}
\usepackage{multirow}
\usepackage{booktabs} 
\usepackage[T1]{fontenc}
\usepackage{algorithm}
\usepackage[noend]{algpseudocode}
\usepackage{caption}
\usepackage{subcaption}

\usepackage{soul} %strike through text using \st{...}

\begin{document}
\title{A Conversational Digital Assistant for Intelligent Process Automation}%\thanks{}}
\titlerunning{Conversational Assistant for IPA}
\author{
Yara Rizk \and
Vatche Isahagian \and
Scott Boag \and
Yasaman Khazaeni \and
Merve Unuvar \and
Vinod Muthusamy \and
Rania Khalaf 
}
\authorrunning{Y. Rizk et al.}
\institute{IBM Research AI, Cambridge, MA, USA}%\\
%\email{@ibm.com} }
%
\maketitle              % typeset the header of the contribution

\begin{abstract} % 15 to 250 words
Robotic process automation (RPA) has emerged as the leading approach to automate tasks in business processes. Moving away from back-end automation, RPA automated the mouse-click on user interfaces; this outside-in approach reduced the overhead of updating legacy software. However, its many shortcomings, namely its lack of accessibility to business users, have prevented its widespread adoption in highly regulated industries. In this work, we explore interactive automation in the form of a conversational digital assistant. It allows business users to interact with and customize their automation solutions through natural language. The framework, which creates such assistants, relies on a multi-agent orchestration model and conversational wrappers for autonomous agents including RPAs. We demonstrate the effectiveness of our proposed approach on a loan approval business process and a travel preapproval business process.
\end{abstract}

\section{Introduction}
% RPAs as the current state-of-the-art in business process automation
Business processes are the backbone of business enterprises and organizations~\cite{weske2012bpm}. A business process is a collection of tasks or activities that must be executed in a certain sequence to achieve a goal. In the era of digital transformation, robotic process automation (RPA) presents a low cost approach to inject automation in business processes. An RPA is developed for tasks that are frequent, repetitive, and error-prone. RPAs learn to execute such tasks in the user interface from humans through demonstration, behavior logs or business process descriptions. Unlike back-end automation approaches, this approach reduces the overhead of adopting automation by operating on top of legacy software. 
 
% Shortcomings of current RPAs
However, highly-regulated industries still require human-in-the-loop automation due to compliance regulations, increased risk, and liability. Unfortunately, the end users are not tech-savvy, making it difficult for them to interact with RPAs and other automation solutions in their current format~\cite{jakob2018barriers}. This lack of accessibility hinders a business user's ability to monitor and customize these solutions~\cite{gao2019automated}. A human-consumable interactive automation, through natural language, would reduce the barrier of entry to business process automation. 

% IDW as a solution to these shortcomings and the future state-of-the-art in business process automation
We propose a framework to build conversational digital assistants, merging the two multi-billion dollar markets of RPA \cite{biscotti2020gartner} and enterprise chatbots\footnote{https://www.businessinsider.com/chatbot-market-stats-trends}. An assistant consists of multiple conversational software bots that automate specific tasks within a business process. It can be used by business users and domain experts who do not possess any programming or software development skills to customize and interact with their business process automation solutions through natural language. The assistant can be adopted in different enterprises depending on the domain of the agents, from banking and finance to retail, customer care and others. 
This interaction would foster trust because it makes the system more transparent to the users, allowing them to gain valuable insights into the system's operations. This trust increases the probability of the business users adopting more automation solutions \cite{muthusamy2018towards}. With the increased interest in trusted business processes in this digital transformation era \cite{rosemann2019trust}, the conversational assistant would serve as a crucial enabler to this paradigm. 

Beyond trusted processes, researchers have investigated business process individualization \cite{wurm2019design}. Pursued by many businesses to differentiate themselves from a large pool of competitors, process individualization has historically been a challenge because it reduces efficiency and increases costs. The digital revolution overcame these challenges by enabling automated modification of individualized processes. Our framework can contribute to this space by further reducing the overhead of creating custom processes and monitoring them through natural language interactions. 

% Proposed framework and its characteristics to realise IDW
The main research question we address in this work is: \textit{what are the necessary characteristics of a software framework that enables interactive automation in business processes through natural language?}
To that end, we present a unified conversational multi-agent orchestration assistant, which consists of multiple building blocks. Skills, including RPAs, automate tasks within a business process They can be composed into more powerful automation bots, and wrapped as conversational agents to interact with business users. As the user converses with the assistant, the orchestrator determines which agents should respond to the user. Therefore, the assistant provides business users access to their RPAs through natural language. We present a taxonomy of skills and agents, define an agent contract to enable such an orchestration and present an orchestration workflow that integrates diverse conversational agents. 

Next, we briefly discuss related work in two main relevant fields: business process automation and conversational agents. Then, we present our proposed framework, before discussing our qualitative analysis on two use cases: travel preapproval, and loan application processing. 

\section{Related Work}
% TODO: add references on RPA collaboration (if any) in 2.1
% TODO: focus on RPA related conversation bots (if any) in 2.2
% TODO: add citations to relevant references from AAAI-IPA workshop
% TODO: add refs from BPM 2019 proceedings (management track): trust ref, ?

\subsection{Business Process Automation}
RPAs have recently been the main driver of the digital transformation with their light-weight approach to automating repetitive tasks \cite{van2018robotic}. They have enabled automation in multiple enterprises including accounting \cite{moffitt2018robotic}, auditing \cite{fernandez2018impacts}, human resources \cite{papageorgiou2018transforming}, banking \cite{stople2017lightweight}, public administration \cite{houy2019robotic} and energy sectors \cite{lacity2015robotic}. Multiple RPA vendors have offered state-of-the-art solutions to clients in various sectors. A survey of these products can be found in \cite{agostinelli2020towards}.  

These RPAs have leveraged diverse technological advancements in the fields of artificial intelligence and software development. Gao et al. adopted deep learning optical character recognition (OCR) and classification models in document flow automation within a debt collector business process \cite{gao2019automated}. RPAs identified relationships between tasks from user behavior in \cite{wroblewska2018robotic} and used first-order logic to deduce automation rules. Crucial to RPA's success is automatically identifying RPA-eligible tasks; \cite{leopold2018identifying} relied on supervised machine learning and natural language processing of business process descriptions to identify these tasks.

Business process automation takes a step beyond RPAs to automate decision making in business processes. Marella et al. identified the field of automated planning as an enabler to more sophisticated business process automation \cite{marrella2017automated}. Machine learning is another enabler; deep learning models, long short-term memory recurrent neural networks specifically, have also been trained to model business processes, a crucial component for more advanced automation \cite{camargo2019learning}. Machine learning algorithms like support vector machines, shallow neural networks and random forests have been adopt in process mining applications \cite{tello2019machine}. An interactive process mining recommender system for business process discovery based on machine learning has also been investigated in the literature \cite{seeliger2019processexplorer}. However, end-to-end process automation has not been widely adopted in enterprises due to business users' lack of trust in such technology \cite{jan2020ai} and limited accessibility and customization capabilities. Since business users lack the technical skills to monitor and customize automation solutions, a natural language interface may be key to the success of the digital transformation. 

\subsection{Conversational Agents} 
Enterprises have been interested in natural language processing advancements, given their heavy reliance on this communication modality. Considering the breadth of this field, we will only focus on conversational agents that are most relevant to the scope of our framework. Enterprise chatbots, a multi-billion dollar market dominated by giants like IBM, Google and Amazon, have been researched for decades and experienced significant improvements recently \cite{galitsky2019developing}. They have evolved from simple question answering customer support bots to fully autonomous assistants capable of performing tasks on behalf of humans \cite{galitsky2019developing}. One shopping bot is capable of custom-pricing products to increase sales \cite{heo2018chatbot}. Food delivery services adopted a delivery bot to reduce the effort customers need to order pizza \cite{heo2018chatbot}. Another bot crowdsourced answers generated by multiple chatbots, gradually reducing the reliance on the crowd through learned selection models \cite{huang2018evorus}. The model estimated the likelihood of an agent returning the correct answer based on the crowd's votes and previous answer selections. Instead, we adopt a different orchestration model that doesn't solely rely on user feedback but has a more sophisticated scoring and selection model to pick one or more agents.  

Researchers have also combined RPAs and chatbots to increase automation. A chatbot for agile software development teams was developed to provide insights into a team's performance by analyzing commits in version control software \cite{matthies2019additional}. From a business process workflow, Kalia et al. derived a dialog tree-based chatbot to converse about the process \cite{kalia2017quark}. Despite the systematic approach to chatbot design, it required significant effort and domain expertise. 

\section{Proposed Framework} 
\begin{figure}[tbp!]
\centering
\begin{subfigure}[t]{0.5\textwidth}
\includegraphics[width=\textwidth]{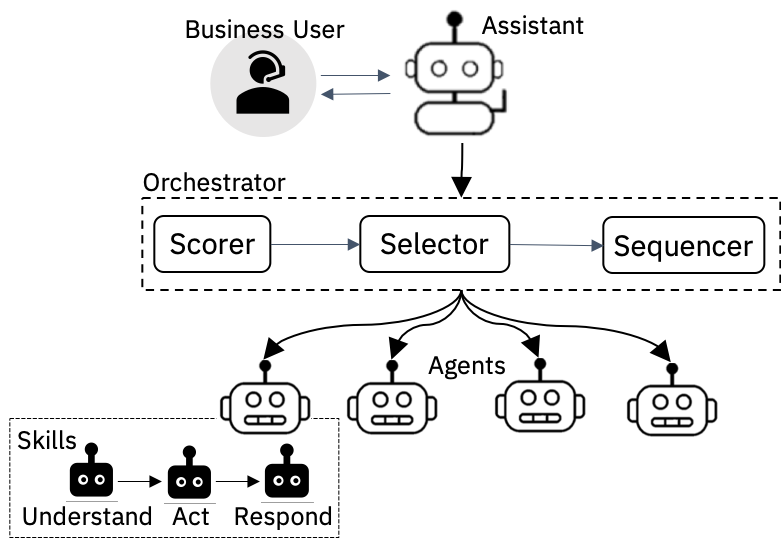}
% \caption{ .}
% \label{fig:}
\end{subfigure}
\hfill
\begin{subfigure}[t]{0.45\textwidth}
\includegraphics[width=\textwidth]{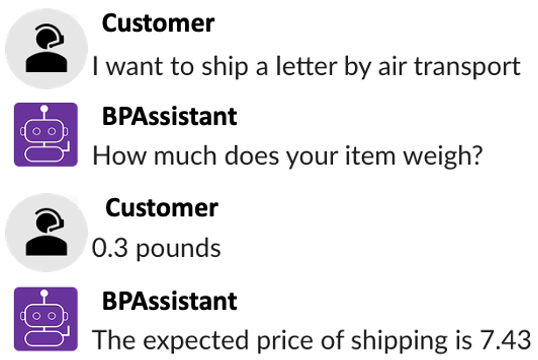}
% \caption{.}
% \label{fig:}
\end{subfigure}
\caption{A Conversational Digital Assistant Framework}
\label{fig:assistant}
\end{figure}

The conversational digital assistant provides business users with two core functionalities beyond RPAs: conversational interaction and collaborative automation. Our proposed framework, illustrated in Fig. \ref{fig:assistant}, achieves these functionalities by relying on an orchestrator capable of coordinating the execution of agents within the system, and agents that can converse with users in natural language while executing tasks in a business process \cite{rizk2020unified}. Agents are composed of skills which can perform natural language understanding and generation in addition to task automation (e.g. RPAs). The orchestrator expects agents to adhere to a specific contract to determine which agents respond to a user's utterance. This modular approach simplifies the task of converting RPA to conversational agents, the system's maintenance and life cycle. It also allows us to add or remove certain domains and functionality from the scope of the assistant with minimal effort.

% \begin{figure}[tb]
%     \centering
%     \includegraphics[width=0.6\linewidth]{Assistant.png}
%     \caption{A Conversational Digital Assistant Framework}
%     \label{fig:assistant}
% \end{figure}

\subsection{Skills}
Skills are the assistant's building blocks and consist of atomic functions to \textit{understand} a user's request, \textit{act} to satisfy the user's request, and \textit{respond} to the user. \textit{Understand} skills are generally natural language understanding functions that determine the user's intent and identify any entities in the user's utterance that would be needed by the \textit{act} skill to properly execute. Such skills can be created using existing tools that create dialog bots such as IBM's Watson Assistant or Google's Dialogflow which can involve defining entities and intents and providing examples of them. 

\textit{Act} skills are generally RPAs that execute the user's intent to produce an outcome. They can be of two types. World-changing skills can change the state of the world, i.e. have side-effects on their environment; examples include skills that send emails or check credit scores. Non-world changing skills do not have side effects on their environment; examples include skills that read emails or check a bank account status. \textit{Act} skills automate business process tasks and can be placed at decision points within a process to move it forward.

\textit{Respond} skills produce a human-consumable response from the \textit{act} skill's output. This can be a natural language utterance, a visual representation, or another modality. This skill can be as simple as adopting a template response or as sophisticated as a deep learning text generation model. 

\subsection{Agents}
In our framework, an agent is simply a conversational RPA: the RPA is preceded by an \textit{understand} skill and succeeded by a \textit{respond} skill. Thus, we obtain an interactive RPA that communicates with business users in natural language. This modular approach enables the integration of RPAs that were not inherently interactive, while reducing the overhead of improving agents throughout their lifetime. However, not all RPAs are suitable to be conversational RPAs.

Composing an agent using the \textit{understand}-\textit{act}-\textit{respond} pipeline achieves the interactive RPA aspect of the assistant and creates a standardized agent creation method. Enforcing a contract enables their integration within the same assistant. Thus, agents could cooperate on tasks and achieve more powerful functionality. Agents receive the input utterance and current context or state; they return a response, an updated context, a confidence in their response (a numerical value between 0 and 1 that quantifies their comprehension of the input and relevance of their response) and possibly other flags related to the conversation. 

Within the conversational digital assistant paradigm, we identified five main types of agents. While many more can be created, we consider the ones that would be most relevant and commonly used in process automation: dialog, information retrieval, task execution, data analytics, and alert agents. 

\noindent{\textbf{Dialog agents}} provide more human-like interactions. They are composed of an \textit{understand} skill and a \textit{respond} skill to answer user queries. They do not change the world and they can be implemented using any conversational agent technology; examples include chit-chat, FAQ or ``help'', etc. 

\noindent{\textbf{Information retrieval agents}} query information sources to achieve their goal. They perform advanced reasoning to respond to user queries or information retrieval tasks in a process. They are composed of an \textit{understand} skill, an information retrieval RPA and a \textit{respond} skill. In general, most agents within this category do not contain world-changing skills (exception: credit score checks). 

\noindent{\textbf{Task execution agents}} perform tasks within a business process that change the state of the world by moving the business process forward. Examples include submitting applications, filling in information in forms, and making decisions at decision points. Such agents may more intuitively fit within the RPA definition. 

\noindent{\textbf{Data analytics agents}}cover a wide scope of functionality from data transformation and modeling to predictions and recommendations. Such agents go beyond information retrieval which provide factual and statistical data but do not go as far as task execution agents that act on the information; they still rely on humans to drive the business process. Examples include visualization and data export agents that manipulate and transform the data to more complex business process forecasting and performance prediction models. 
 
\noindent{\textbf{Alerting agents}} allow users to conversationally customize alerts and notifications triggered by the occurrence of specific events within or related to the process, in essence enabling asynchronous monitoring of the process \cite{rizk2020snooze}. %Such agents consist of an \textit{understand} skill that allows users to set up and customize their alerts, an \textit{act} skill that creates a watcher to trigger alerts, and a \textit{respond} skill that creates the notification or response when setting up the alerts. 

\subsection{Orchestration}
\label{sec:orchestration}
The orchestrator is the core component that allows interactive and cooperating agents in our framework. Its main functionalities include selecting a subset of agents that must respond to a user's request, managing the context and passing it among agents, and acting as the central dialog manager (controls a multi-turn conversation's state and flow).   

Selecting the agent(s) to respond to a user's request can be achieved by different orchestrators. Stateless orchestrators, unlike their stateful counterparts, do not possess a central state tracker, i.e. context is maintained in every turn by passing context variables between the orchestrator and the agents. Maintaining context about the dialog also influences agent selection: if the user is in the middle a conversation with an agent, the orchesetrator must ensure a smooth interaction. If users digress from the conversation (i.e. move away before completing the conversation) or provide ambiguous statements, the dialog manager should properly handle such situations. Apriori orchestrators select agents based on the user input and some knowledge about agent capabilities. On the other hand, posterior orchestrators request a preview response from agents that factors into the orchestrator's selection. Such orchestrators could also request a confidence score to better assess an agent's response. Confidence is a quantification of an agent's understanding of the input and relevance of its response to it. Posterior orchestrators require agents to have a preview mode in order not to change the world if they are not selected for execution. They are agnostic to what agents exist in the assistant as long as they abide by the contract; they simply need to know of the agents' existence and how to reach them (API endpoint).

Each orchestration model has specific computational costs, architectural requirements, and contract constraints. These factors influence the assistant's orchestration model. In this work, we adopt a stateless, posterior orchestrator, called 3S orchestrator, that consists of three main steps: scoring, selecting, and sequencing \cite{yurochkin2019online,upadhyay2019bandit}. Outlined in Algorithm \ref{alg:orch3S}, it assumes an agent ($a_i$) contract that returns a preview response ($r_i$), a confidence in the response ($c_i$), and a stickiness value ($\kappa_i$) that indicates whether the agent has been interacting with the user in previous turns (i.e. is in the middle of a conversation). A preview mode ensures that world-changing agents do not cause irreversible changes before the orchestrator makes its selection. Furthermore, preview and execute modes can be adopted to optimize the computation/latency of non-world changing agents. The nomenclature that is adopted throughout this work is defined in Table \ref{tab:nomenclature}.

% \begin{table}[tb]
%     \centering
%     \begin{tabular}{c|l}
%         \toprule
%         Symbol &  Definition \\
%         \midrule
%         $A$ & Set of agents \\
%         $n = |A| $ & Number of agents \\
%         $a_i \in A$ & Agent $i$ \\
%         $c_i \in [0,1]$ & Confidence of agent $a_i$ \\
%         $\kappa_i \in \{0,1\}$ & Stickiness of agent $a_i$ \\
%         $r_i$ & Response of agent $a_i$ \\
%         $u$ & User's utterance \\
%         $f_i(.)$ & Function per agent $a_i$ \\
%         $s_i$ & Score for agent $a_i$ \\
%         $g(.)$ & Scorer function, computes a score per agent $a_i$ \\
%         $A_s$ & Set of selected agents \\
%         $k = |A_s|$ & Number of selected agents \\
%         $R$ & Final Response \\
%         \bottomrule
%     \end{tabular}
%     \caption{Nomenclature}
%     \label{tab:nomenclature}
% \end{table}

\begin{table}[tb]
    \centering
    \begin{tabular}{c l| c l}
        \toprule
        Symbol &  Definition & Symbol &  Definition\\
        \midrule
        $A$ & Set of agents & $n = |A| $ & Number of agents \\
        $a_i \in A$ & Agent $i$ & $r_i$ & Response of agent $a_i$ \\
        $c_i \in [0,1]$ & Agent $a_i$'s confidence & $\kappa_i \in \{0,1\}$ & Agent $a_i$'s stickiness\\
        $u$ & User's utterance & $f_i(.)$ & Function per agent $a_i$ \\
        $s_i$ & Agent $a_i$'s score & $g(.)$ & Scorer function, computes a score per agent $a_i$ \\
        $A_s$ & Set of selected agents & $h(.)$ & Selector function \\
        $R$ & Final response & $k = |A_s|$ & Number of selected agents \\
        \bottomrule
    \end{tabular}
    \caption{Nomenclature}
    \label{tab:nomenclature}
\end{table}

\begin{algorithm}[tb]
\caption{3S Orchestration}\label{alg:orch3S}
\begin{algorithmic}[1]
\Procedure{Broadcaster}{$u$, $A$} 
\For{$a_i \in A = \{a_1, ... , a_n\}$ }
\State $(c_i, \kappa_i, r_i) = f_i({u})$
\EndFor
\EndProcedure
\Procedure{Orchestrator}{Responses, Confidences} 
\State Scorer: $s_i = g(c_i, \kappa_i), \forall a_i \in A$
\State Selector: $A_s = h(\{s_i \forall a_i \in A \})$ 
\State Sequencer: $R = order(\{r_i, \forall a_i \in A_s\})$
\State \textbf{return} $R$
\EndProcedure
\end{algorithmic}
\end{algorithm}

\begin{figure}[t]
    \centering
    \includegraphics[width=0.99\linewidth]{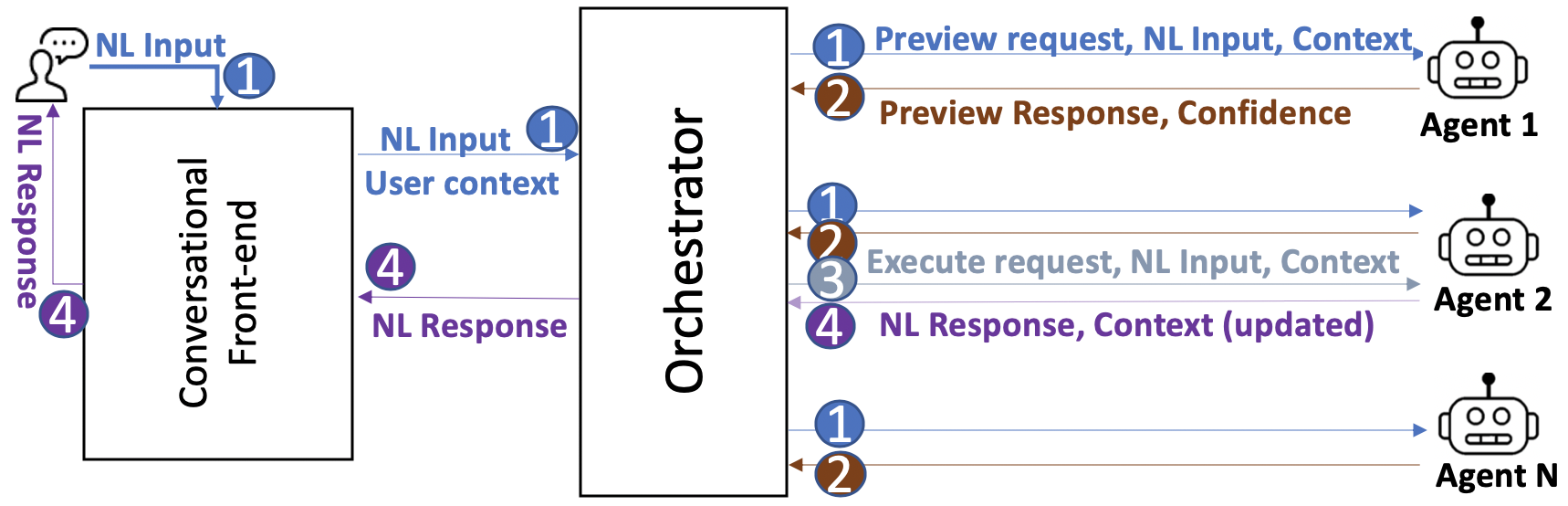}
    \caption{Graphical Representation of a Posterior Orchestration Pipeline}
    \label{fig:orch_flow}
\end{figure}

\noindent{\textbf{Scorer.}} When the orchestrator receives a natural language utterance from the user, as shown in Fig. \ref{fig:orch_flow}, it forwards the input to all agents. Once it receives the agents' preview responses, the scorer processes the agent confidences (and possibly other variables) to obtain normalized values. This is necessary in multi-agent environments where agents are developed by independent software developers or diverse frameworks that compute confidences differently. A scorer can be as simple as an identity function that does not modify the confidences (if inherently normalized) or as complex as a Bayesian approach that incorporates statistical and probabilistic models to scale the confidences. For example, per our agents' contract, a confidence between 0 and 1 is returned in addition to a flag, called stickiness equal to 0 or 1, that states whether an agent had been selected in the previous turn and is expecting an answer from the user. One scorer we adopted simply takes the maximum of both values, $s_i = max(c_i,\kappa_i)$.

\noindent{\textbf{Selector.}} Next, the selector processes the scores and determines which agent(s) must execute to respond to the user. The selector model can be a simple \textit{Top 1 (or Top K)} selector that picks the agent (or K agents) with the highest score (above a minimum threshold, $T$, to identify cases outside the scope of the assistant), e.g. $A_s = max \{s_i : a_i \in A \quad \mathrm{s.t.} \quad s_i > T$\} (adopted in our experiments). It could also be a machine learning algorithm that utilizes other features such as previous conversation turns to select the next best agent(s). In a supervised learning domain, labels can represent agents that are the output of classifiers and the input to the classifier would be a feature vector consisting of the agents' scores and/or user utterances. In a reinforcement learning domain, actions can represent agents and the environment produces a reward when the correct agent is selected. With enough data, deep learning models can be trained.

\noindent{\textbf{Sequencer.}} If multiple agents were selected for execution, a sequencer determines the order to execute these agents and show their responses' to the user. This is crucial to the proper execution of collaborating agents since one agent's output is another's input; agents' execution is not independent. Various approaches ranging from relatively simple heuristic rules to more complex planning-based algorithms can be adopted.

\section{Empirical Results}

\subsection{Implementation and Use Cases}
\begin{figure}[tb]
    \centering
    \includegraphics[width=0.99\linewidth]{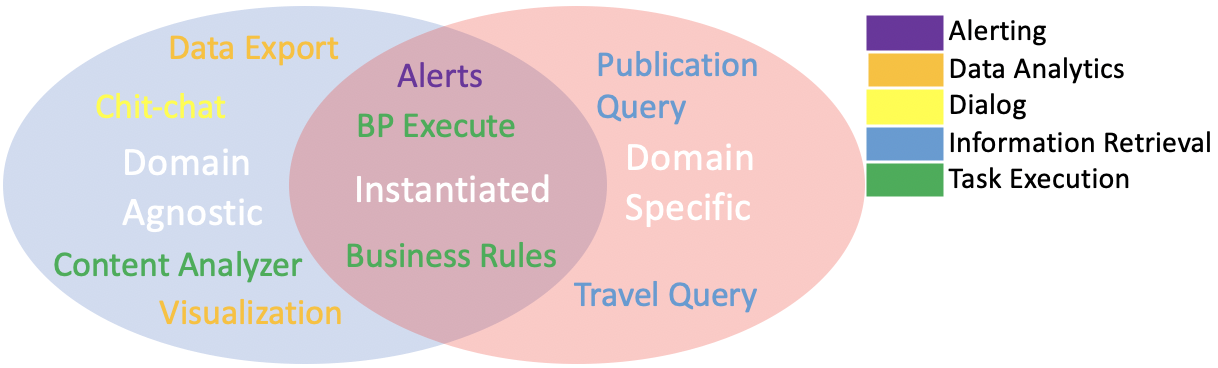}
    \caption{Agents in \textit{BPAssistant}}
    \label{fig:BPAssistant_Agents}
\end{figure}

The proposed framework is implemented in Python where most skills are accessible through API endpoints. A \textit{top K} orchestrator is exposed as a Rest API, and called by a Slack bot\footnote{https://api.slack.com/bot-users\#bots-overview}. Conversational skills were implemented in Watson Assistant\footnote{https://www.ibm.com/cloud/watson-assistant}. \textit{Act} skills were either external microservices or internal Python functions. The assistant, referred to as \textit{BPAssistant}, consists of multiple agents (see Fig. \ref{fig:BPAssistant_Agents}) to handle two simplified business processes: loan application and travel preapproval \cite{rizk2020unified}. Some agents are domain agnostic and can operate on multiple business processes without any overhead. Others are domain specific, developed for a very specific task in one of the business process. Finally, a set of agents were created from domain agnostic agents but must be instantiated for a specific domain. Hence, they require some configuration when deployed in a process. 

\begin{figure}[tb]
    \centering
    \includegraphics[width=0.99\linewidth]{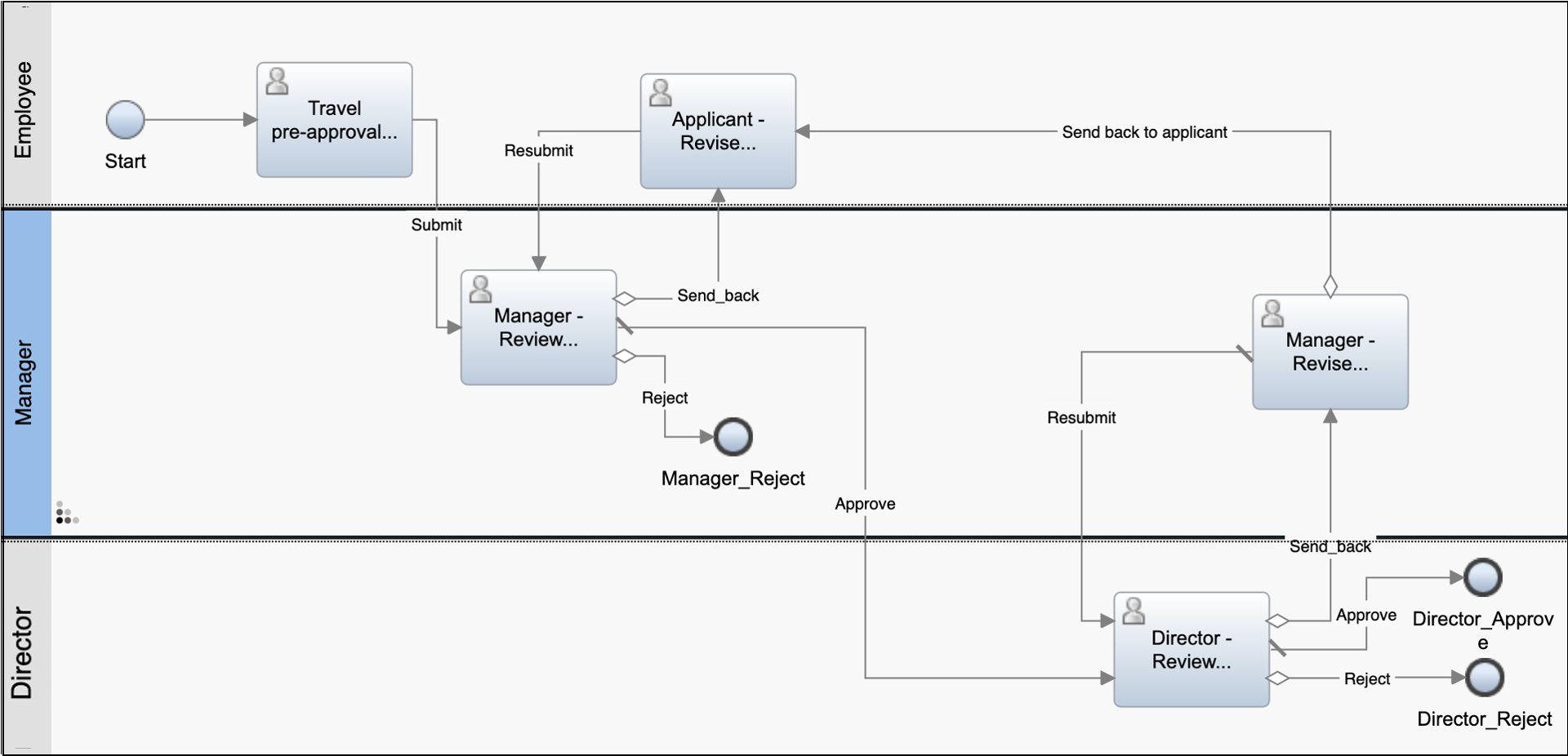}
    \caption{Travel Preapproval Process}
    \label{fig:bpm_travel}
\end{figure}

The travel use case considers two persons that can interact with the \textit{BPAssistant}: employees and managers. The travel preapproval process, shown in Fig. \ref{fig:bpm_travel}, was implemented in a business process management software. Employees submit a travel preapproval request to attend a conference (or event) by filling a form. Once submitted, the employee's manager processes the application. If approved by the manager, the application is forwarded to the director who makes the final decision on the travel request. 

The simplified loan application process consists of a bank customer submitting a loan application and a bank officer processing this request to determine whether to approve or reject it. The assistant can help a loan officer automate certain parts of the process, in addition to query and analyze the process data.

These use cases are interesting because of their relevance to many enterprises, especially those that are about to start or are in the middle of their digital transformation journey. These processes (or parts of them) may need to suddenly become agile overnight in response to a pandemic (e.g. loan officer remotely approving a loan), or readjusted processes to address a sudden change in company policies (no flights to certain countries). Creating interactive agents from RPAs that automate tasks would enable employers' adoption of such systems. Natural language interactions reduce the necessary learning curve, leading to quick deployment. 

\subsection{Conversations with \textit{BPAssistant}}
\begin{figure}[tbp!]
\centering
\begin{subfigure}[t]{0.48\textwidth}
\includegraphics[width=\textwidth]{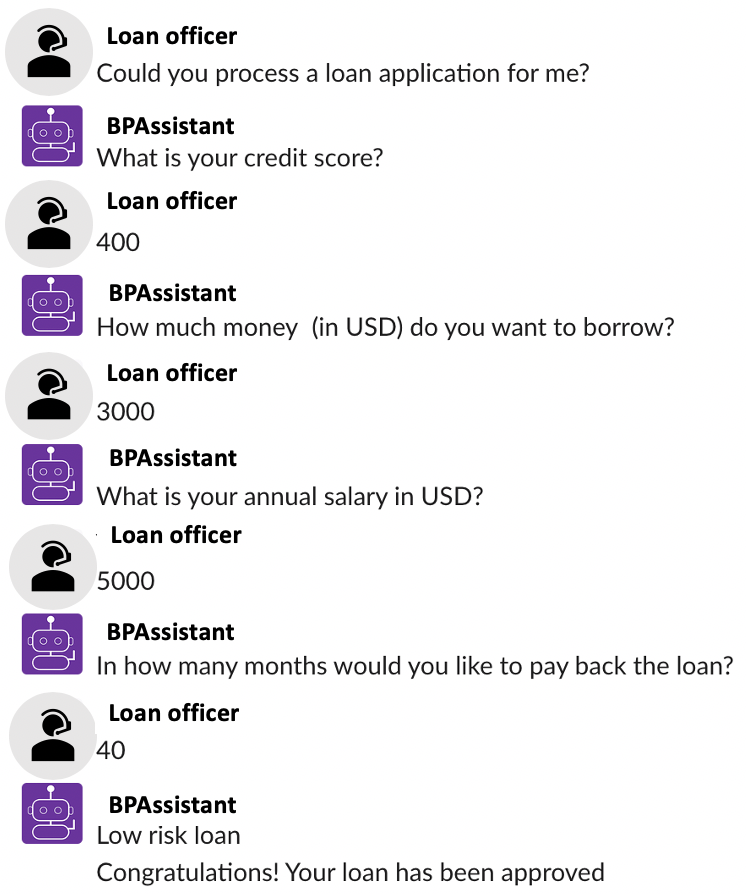}
\caption{Conversation driven.}
\label{fig:business_rules_WA}
\end{subfigure}
\hfill
\begin{subfigure}[t]{0.48\textwidth}
\includegraphics[width=\textwidth]{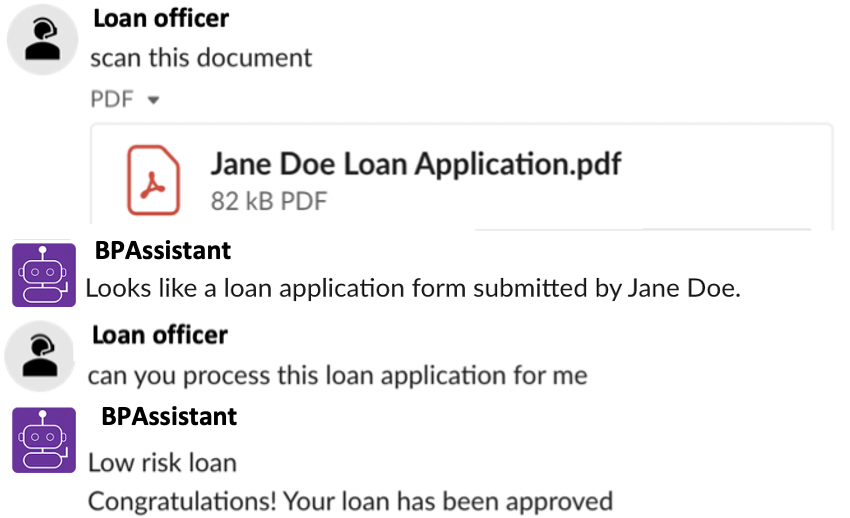}
\caption{RPA cooperation.}
\label{fig:baca2odm}
\end{subfigure}
\caption{Conversation with \textit{BPAssistant}: business rules agent.}
\label{fig:business_rules_agent}
\end{figure}

\textit{BPAssistant} allows users to accomplish different goals by relaying their intentions in natural language. First, consider a loan officer who wants to process a customer's loan application while consulting the bank's rule engine. The ``Business Rules'' agent would allow the loan officer to make decisions about tasks in a business process such as approving or rejecting the loan application. Fig. \ref{fig:business_rules_WA} shows a sample multi-turn conversation with this agent through the assistant. The agent was instantiated for the banking domain and implemented in IBM Decision Service\footnote{https://www.ibm.com/products/operational-decision-manager}, where users input the available information and the agent recommends whether to approve or reject the loan based on the business rules. %The Decision Service is domain agnostic and rules can be defined for any domain.%; in this case, we instantiated it for the banking domain and included it as an agent in \textit{BPAssistant}. 

% \begin{figure}[tb]
%     \centering
%     \includegraphics[width=0.6\linewidth]{business_rules_2.png}
%     \caption{Conversation with \textit{BPAssistant}: business rules agent.}
%     \label{fig:business_rules}
% \end{figure}

However, the interaction can be improved by automating information gathering using an RPA (configured as an agent) to extract this information from a document, for example. 
% Example: BACA --> ODM
Fig. \ref{fig:baca2odm} shows the same example but with cooperation among agents. The ``content analyzer'' agent, capable of extracting information from PDF documents, retrieves key-value pairs from the loan application file and saves them in the context of the orchestrator. Then, the ``business rules'' agent obtains this information through the orchestrator to determine whether the loan should be approved or rejected. Outside of the assistant framework, enabling such an RPA cooperation would require the creation of another RPA that transfers the data from the first RPA (content analyzer) to the second RPA (business engine). The orchestrator presents a more generalizable paradigm to achieve this functionality. 

% \begin{figure}[tb]
%     \centering
%     \includegraphics[width=0.6\linewidth]{BACA2ODM_2.png}
%     \caption{Conversation with \textit{BPAssistant}: RPA cooperation.}
%     \label{fig:baca2odm}
% \end{figure}

\begin{figure}[tb]
    \centering
    \includegraphics[width=0.6\linewidth]{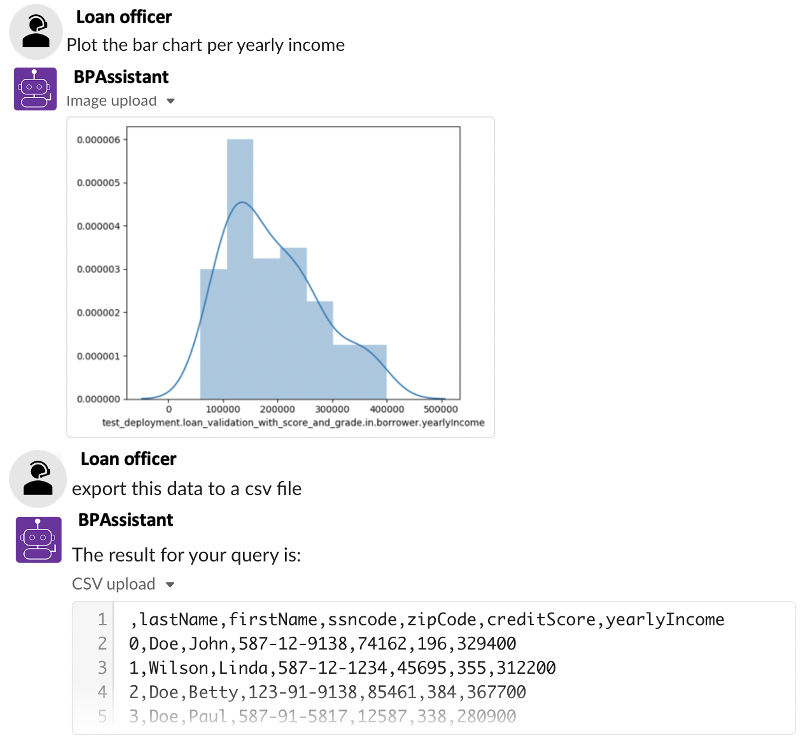}
    \caption{Multi-modal responses from the \textit{BPAssistant}.}
    \label{fig:loan_convo}
\end{figure}

The loan officer can also use \textit{BPAssistant} to analyze recent loan applications by querying the data associated with the loan process, as shown in Table \ref{tab:bpBot_convos}. The natural language utterances in the table invoke the ``Business Process Query'' agent capable of converting natural language sentences to formal queries that are executed on a database to retrieve the data \cite{sen2019natural}. Furthermore, the loan officer can manipulate the data by plotting it and exporting it to a file, using the Visualization and Data Export agents, respectively. Both agents use the data from information retrieval agents, stored in the context of the orchestrator, as input to fulfill the user's request. Fig. \ref{fig:loan_convo} displays these multi-modal responses, namely the visualization and CSV file generated based on the queried data.%, upon the loan officer's request for plots. 

Now, let's say the loan officer needs to attend training or a seminar on new loan processing procedures at the bank's headquarters. The travel costs can be expensed to the bank since it is a work trip\footnote{Travel preapproval processes are common in many enterprises including IBM Research. Employees submit preapprovals to attend academic conferences, client meetings, training events, etc.}. The ``Business Process (BP) Execute'' agent can move the business process forward by asking the assistant to execute a step in the process on their behalf, such as submitting a travel request by saying ``submit a travel request to the headquarters''. \textit{BPAssistant} would then submit a request on the employee's behalf and automatically populate the fields of the form by using appropriate \textit{act} skills that can retrieve information. 

Furthermore, a bank manager can use the same \textit{BPAssistant} to set up alerts about employees' submitted travel requests using the ``Alerts'' agent. This agent enables users to create and customize alerts related to a process \cite{rizk2020snooze}. Once an notification is received, the manager can review the employee's travel preapproval request and determine whether to approve it or not, as shown in Table \ref{tab:bpBot_convos}. 

\begin{table}[tb]\small
    \centering
    \begin{tabular}{p{9cm} c}
         \textbf{Conversation} & \textbf{Responding agent} \\
         \toprule
         Loan Officer: Who are the top 3 borrowers with average amount more than 10000 & \multirow{2}{*}{Data Query}\\
         \textit{BPAssistant}: These are the average value: 1). 584,917\$ for J. Smith, 2). 575,692\$ for V. Doe, 3). 557,615\$ for Y. Doe & \\
         \midrule
         Loan Officer: List all borrowers with yearly income more than 50000 but credit score less than 150 & \multirow{2}{*}{Data Query}\\
         \textit{BPAssistant}: Total records found are 82. Here is the link: $<$url$>$ & \\
         \midrule
         Loan Officer: Plot the bar chart per yearly income
 & \multirow{2}{*}{Visualization}\\
         \textit{BPAssistant}: $<$image$>$ (Fig. \ref{fig:loan_convo}) & \\
         \midrule
         Loan Officer: Export this data to a CSV file & \multirow{2}{*}{Data Export}\\
         \textit{BPAssistant}: The result for your query is: $<$csv$>$ file (Fig. \ref{fig:loan_convo}) & \\
         \midrule
         \midrule
         Manager: Hello & \multirow{2}{*}{Chit-Chat}\\
         \textit{BPAssistant}: Hi there & {} \\
         \midrule
         Manager: How many travel requests does John Smith have? & Travel Query \\
         \textit{BPAssistant}: John Smith has 1 application \\
         \midrule
         Manager: Approve John Smith's request & Business Process \\
         \textit{BPAssistant}: John Smith's application has been approved & Task Execution \\
         \bottomrule
    \end{tabular}
    \caption{Sample conversation with \textit{BPAssistant}}
    \label{tab:bpBot_convos}
\end{table}

\subsection{Discussion}
Based on our prototypical implementation above, we observed multiple advantages and some limitations of the proposed framework. First, the assistant successfully handled the users' diverse requests by orchestrating agents from multiple domains. Business users interacted with RPAs such as the ``alerts'' agent or the ``business rules'' agent through natural language without switching between multiple interfaces. Querying their data did not require business users to possess the knowledge of formulating formal queries executed on their databases; they simply formulated natural language statements using business domain terminology that they frequently use. Performing tasks within their business processes no longer required them to juggle multiple user interfaces either. 

The framework eliminated the need to create custom RPAs by the business user; instead, developers created general RPAs that could be customized through natural language by the business user in the assistant. However, this shift came at the cost of creating conversational wrappers for the RPAs. Future iterations should consider approaches to automate this task to further reduce the coding overhead. Additionally, automated composition techniques, such as the ones explored in \cite{chakraborti2020d3ba,muise2019planning}, can further reduce the overhead of authoring these agents by automatically composing sophisticated agents out of more atomic skills.

As more heterogeneous agents are added to the assistant, the orchestration problem becomes more complex, especially when considering posterior orchestration approaches. One method to reduce the cost of previewing agent responses is to implement a hybrid orchestration method that adopts an apriori algorithm to select a subset of agents to ask for a preview. This reduces the computational cost as the system is scaled, while maintaining high selection accuracy.

The stateless orchestrator offered a lightweight solution to enable cooperation among agents, the second contribution of this work. However, as the size of data exchanged between agents increases, a stateful implementation may be more suitable. Shareable data can be stored in memory while the orchestrator keeps track of other state-related information to handle more complex dialog constructs and cooperation opportunities. A stateful orchestrator that combines apriori and posterior models could also alleviate the problem of legacy programs not having a preview mode to invoke in posterior orchestrators. The statefulness of the orchestrator would compensate for the missing preview responses by using other features of the system to make the best selection possible. 

Furthermore, the orchestrator's modular principles simplify the problem of system lifecycle. As business processes become obsolete, the agents linked to them can simply be deactivated without incurring costs to update the framework or other agents. The framework also supports agent versions: as updates to agents are rolled out, the affected agents can be independently updated. 

We believe this framework generalizes well to many types of RPAs because the goal of RPAs is to reduce the amount of work done by humans. Hence, they require minimal human intervention beyond launching the RPAs which can be implemented as a natural language instruction. Functions that still require a large amount of input from users may not be suitable for conversational interaction and should remain in their original user interface. However, as the number of RPAs increases, users will have too many RPAs to keep track of individually; we believe a unified, conversational interface would simplify the life of users. 

In summary, our framework addressed some of the key weaknesses of RPAs, namely providing an interactive, human-in-the-loop automation assistant and cooperative RPAs. Even though the framework can undergo further improvements, it is a solid step forward towards interactive business process automation that can gain business users' trust in process automation solutions and advance service lines like customer care and process automation. 
\section{Conclusion}
Conversational automation solutions will be critical to the digital transformation. To that end, we presented a framework that combines RPAs with conversational agents (or chatbots), both popular paradigms in business enterprises, to create an interactive business process automation solution. The framework relies on multi-agent orchestration where conversational agents are composed from RPA skills. The resulting assistant allows business users to monitor and customize their business process automation solutions through natural language. Future work will incorporate more sophisticated orchestration models and autonomous agents to address existing challenges in the current framework. 

\section*{Acknowledgement}
The authors would like to acknowledge Tathagata Chakraborti, Pierre Feillet, and Stephane Mery for the valuable conversations that contributed to expanding the vision of this work. 

\bibliographystyle{splncs04}
\bibliography{refs}

\begin{thebibliography}{10}
\providecommand{\url}[1]{\texttt{#1}}
\providecommand{\urlprefix}{URL }
\providecommand{\doi}[1]{https://doi.org/#1}

\bibitem{van2018robotic}
van~der Aalst, W.M., Bichler, M., Heinzl, A.: Robotic process automation (2018)

\bibitem{agostinelli2020towards}
Agostinelli, S., Marrella, A., Mecella, M.: Towards intelligent robotic process
  automation for bpmers. arXiv preprint arXiv:2001.00804  (2020)

\bibitem{biscotti2020gartner}
Biscotti, F., Mehta, V., Villa, A., Bhullar, B., Tornbohm, C.: Market share
  analysis: Robotic process automation, worldwide, 2019. Tech. rep. (2020)

\bibitem{camargo2019learning}
Camargo, M., Dumas, M., Gonz{\'a}lez-Rojas, O.: Learning accurate lstm models
  of business processes. In: International Conference on Business Process
  Management. pp. 286--302. Springer (2019)

\bibitem{chakraborti2020d3ba}
Chakraborti, T., Khazaeni, Y.: {D3BA: A Tool for Optimizing Business Processes
  Using Non-Deterministic Planning}. In: AAAI IPA (2020)

\bibitem{fernandez2018impacts}
Fernandez, D., Aman, A.: Impacts of robotic process automation on global
  accounting services. Asian Journal of Accounting and Governance  \textbf{9},
  123--132 (2018)

\bibitem{galitsky2019developing}
Galitsky, B.: Developing Enterprise Chatbots. Springer (2019)

\bibitem{gao2019automated}
Gao, J., van Zelst, S.J., Lu, X., van~der Aalst, W.M.: Automated robotic
  process automation: A self-learning approach. In: OTM Confederated Int. Conf.
  On the Move to Meaningful Internet Systems. pp. 95--112. Springer (2019)

\bibitem{heo2018chatbot}
Heo, M., et~al.: Chatbot as a new business communication tool: The case of
  naver talktalk. Business Communication Research and Practice  \textbf{1}(1),
  41--45 (2018)

\bibitem{houy2019robotic}
Houy, C., Hamberg, M., Fettke, P.: Robotic process automation in public
  administrations. Digitalisierung von Staat und Verwaltung  (2019)

\bibitem{huang2018evorus}
Huang, T.H., Chang, J.C., Bigham, J.P.: Evorus: A crowd-powered conversational
  assistant built to automate itself over time. In: Proceedings of the 2018 CHI
  Conference on Human Factors in Computing Systems. pp. 1--13 (2018)

\bibitem{jakob2018barriers}
Jakob, M., Krcmar, H.: Which barriers hinder a successful digital
  transformation in small and medium-sized municipalities in a federal system?
  In: Central and Eastern European eDem and eGov Days. pp. 141--150 (2018)

\bibitem{jan2020ai}
Jan, S.T., Ishakian, V., Muthusamy, V.: Ai trust in business processes: The
  need for process-aware explanations. IAAI 2020  (2020)

\bibitem{kalia2017quark}
Kalia, A.K., Telang, P.R., Xiao, J., Vukovic, M.: Quark: a methodology to
  transform people-driven processes to chatbot services. In: International
  Conference on Service-Oriented Computing. pp. 53--61. Springer (2017)

\bibitem{lacity2015robotic}
Lacity, M., Willcocks, L.P., Craig, A.: Robotic process automation: mature
  capabilities in the energy sector. Technical Report  (2015)

\bibitem{leopold2018identifying}
Leopold, H., van~der Aa, H., Reijers, H.A.: Identifying candidate tasks for
  robotic process automation in textual process descriptions. In: Enterprise,
  Business-Process and Information Systems Modeling, pp. 67--81. Springer
  (2018)

\bibitem{marrella2017automated}
Marrella, A.: What automated planning can do for business process management.
  In: Int. Conf. Business Process Management. pp. 7--19. Springer (2017)

\bibitem{matthies2019additional}
Matthies, C., Dobrigkeit, F., Hesse, G.: An additional set of (automated) eyes:
  chatbots for agile retrospectives. In: Proceedings of the 1st International
  Workshop on Bots in Software Engineering. pp. 34--37. IEEE Press (2019)

\bibitem{moffitt2018robotic}
Moffitt, K.C., Rozario, A.M., Vasarhelyi, M.A.: Robotic process automation for
  auditing. Journal of Emerging Technologies in Accounting  \textbf{15}(1),
  1--10 (2018)

\bibitem{muise2019planning}
Muise, C., Chakraborti, T., Agarwal, S., Bajgar, O., Chaudhary, A.,
  Lastras-Montano, L.A., Ondrej, J., Vodolan, M., Wiecha, C.: Planning for
  goal-oriented dialogue systems. arXiv preprint arXiv:1910.08137  (2019)

\bibitem{muthusamy2018towards}
Muthusamy, V., Slominski, A., Ishakian, V.: Towards enterprise-ready ai
  deployments minimizing the risk of consuming ai models in business
  applications. In: 2018 First International Conference on Artificial
  Intelligence for Industries (AI4I). pp. 108--109. IEEE (2018)

\bibitem{papageorgiou2018transforming}
Papageorgiou, D.: Transforming the hr function through robotic process
  automation. Benefits Quarterly  \textbf{34}(2),  27--30 (2018)

\bibitem{rizk2020unified}
Rizk, Y., Bhandwalder, A., Boag, S., Chakraborti, T., Isahagian, V., Khazaeni,
  Y., Pollok, F., Unuvar, M.: A unified conversational assistant framework for
  business process automation. In: AAAI IPA (2020)

\bibitem{rizk2020snooze}
Rizk, Y., Isahagian, V., Unuvar, M., Khazaeni, Y.: A snooze-less user-aware
  notification system for proactive conversational agents. Intelligent User
  Interfaces Workshop on User-Aware Conversational Agents  (2020)

\bibitem{rosemann2019trust}
Rosemann, M.: Trust-aware process design. In: International Conference on
  Business Process Management. pp. 305--321. Springer (2019)

\bibitem{seeliger2019processexplorer}
Seeliger, A., Guinea, A.S., Nolle, T., M{\"u}hlh{\"a}user, M.: Processexplorer:
  intelligent process mining guidance. In: International Conference on Business
  Process Management. pp. 216--231. Springer (2019)

\bibitem{sen2019natural}
Sen, J., Ozcan, F., Quamar, A., Stager, G., Mittal, A., Jammi, M., Lei, C.,
  Saha, D., Sankaranarayanan, K.: Natural language querying of complex business
  intelligence queries. In: Proceedings of the 2019 International Conference on
  Management of Data. pp. 1997--2000. ACM (2019)

\bibitem{stople2017lightweight}
Stople, A., Steinsund, H., Iden, J., Bygstad, B.: Lightweight it and the it
  function: experiences from robotic process automation in a norwegian bank.
  Bibsys Open Journal Systems  \textbf{25}(1) (2017)

\bibitem{tello2019machine}
Tello, G., Gianini, G., Mizouni, R., Damiani, E.: Machine learning-based
  framework for log-lifting in business process mining applications. In:
  International Conference on Business Process Management. pp. 232--249.
  Springer (2019)

\bibitem{upadhyay2019bandit}
Upadhyay, S., Agarwal, M., Bounneffouf, D., Khazaeni, Y.: A bandit approach to
  posterior dialog orchestration under a budget. NeurIPS Conversational AI
  Workshop  (2019)

\bibitem{weske2012bpm}
Weske, M.: Business Process Management (2012)

\bibitem{wroblewska2018robotic}
Wr{\'o}blewska, A., Stanis{\l}awek, T., Prus-Zaj{\k{a}}czkowski, B., Garncarek,
  {\L}.: Robotic process automation of unstructured data with machine learning.
  Annals of Computer Science and Information Systems  \textbf{16},  9--16
  (2018)

\bibitem{wurm2019design}
Wurm, B., Goel, K., Bandara, W., Rosemann, M.: Design patterns for business
  process individualization. In: International Conference on Business Process
  Management. pp. 370--385. Springer (2019)

\bibitem{yurochkin2019online}
Yurochkin, M., Upadhyay, S., Bouneffouf, D., Agarwal, M., Khazaeni, Y.: Online
  semi-supervised learning with bandit feedback. LLD Workshop at ICLR  (2019)

\end{thebibliography}

\end{document}